\documentclass[conference]{IEEEtran}
\IEEEoverridecommandlockouts
\usepackage{cite}

\usepackage{xcolor}
\usepackage[colorlinks=true, linkcolor=red, urlcolor=red]{hyperref}
\usepackage{orcidlink}

\usepackage{amsmath,amssymb,amsfonts}
\usepackage{algorithmic}
\usepackage{graphicx}
\usepackage{textcomp}
\usepackage{xcolor}
\def\BibTeX{{\rm B\kern-.05em{\sc i\kern-.025em b}\kern-.08em
    T\kern-.1667em\lower.7ex\hbox{E}\kern-.125emX}}

\begin{document}

\title{Real-Time Machine Learning for Embedded Anomaly Detection}

\author{\IEEEauthorblockN{1\textsuperscript{st} Abdelmadjid Benmachiche\orcidlink{0000-0002-0690-2625}}
\IEEEauthorblockA{\textit{Department of Computer Science} \\
\textit{LIMA Laboratory}\\
\textit{Chadli Bendjedid University}\\
El-Tarf, PB 73, 36000, Algeria\\
benmachiche-abdelmadjid@univ-eltarf.dz}

\and
\IEEEauthorblockN{2\textsuperscript{nd} Khadija Rais\orcidlink{0009-0004-3907-7782}}
\IEEEauthorblockA{\textit{Informatics and Systems (LAMIS)} \\
\textit{Echahid Cheikh Larbi Tebessi University}\\
Tebessa, 12002, Algeria \\
khadija.rais@univ-tebessa.dz}

\and
\IEEEauthorblockN{3\textsuperscript{rd} Hamda Slimi\orcidlink{0000-0002-8494-6551}}
\IEEEauthorblockA{\textit{Informatics and Systems (LAMIS)} \\
\textit{Echahid Cheikh Larbi Tebessi University}\\
Tebessa, 12002, Algeria \\
slimi.hamda@univ-tebessa.dz}
}

\maketitle

 \IEEEpubid{%
  \begin{minipage}{\textwidth}
    \vspace{7em}
    \makebox[0pt][l]{\textbf{The second National Conference on Artificial Intelligence and Information Technologies}-\textit{ Chadli Bendjedid El-Tarf University - EL TARF (Algérie)}
     }
  \end{minipage}%
}

\begin{abstract}
The spread of a resource-constrained Internet of Things (IoT) environment and embedded devices has put pressure on the real-time detection of anomalies occurring at the edge. This survey presents an overview of machine-learning methods aimed specifically at on-device anomaly detection with extremely strict constraints for latency, memory, and power consumption. Lightweight algorithms such as Isolation Forest, One-Class SVM,  recurrent architectures, and statistical techniques are compared here according to the realities of embedded implementation. Our survey brings out significant trade-offs of accuracy and computational efficiency of detection, as well as how hardware constraints end up fundamentally redefining algorithm choice. The survey is completed with a set of practical recommendations on the choice of the algorithm depending on the equipment profiles and new trends in TinyML, which can help close the gap between detection capabilities and embedded reality. The paper serves as a strategic roadmap for engineers deploying anomaly detection in edge environments that are constrained by bandwidth and may be safety-critical.
\end{abstract}


\section{Introduction}
An immediate demand has arisen with the fast adoption of IoT and embedded systems across the critical infrastructure, both in industrial control systems and in smart cities, for real-time, on-the-edge anomaly detection. Sending raw sensor data to the cloud to be analyzed is challenging because of bandwidth issues, latency, and privacy issues, which demand edge-based intelligence \cite{adhikari2024recent}.  Here, anomaly detectors need to be extremely efficient on very constrained resources such as small memory (usually less than 100 KB), low-power processors, and hard real-time performance requirements, excluding many traditional machine learning methods.

The past few years have been characterized by runaway research efforts on how to fit anomaly detection methods into these limited settings. This urgency is amplified by the growing vulnerability of IoT environments to cyber-attacks exploiting system weaknesses and user awareness gaps\cite{sedraoui2024intrusion}. In a survey by Adhikari et al. scan across the wide field of IoT anomaly detection, but it is often the case that many suggested solutions are not viable in real embedded settings, as they are too computationally consuming \cite{adhikari2024recent}. The TinyML movement has addressed this gap to some degree by implementing models that are very lightweight and runnable on microcontrollers. Antonini et al. showed a completely unsupervised and flexible anomaly detection system running on a low-cost IoT retrofitting kit, which demonstrates the functionality of on-device learning \cite{antonini2023adaptable}.

Moreover, the issue of concept drift, i.e., the definition of what normal behavior is, changes over time, and presents a major difficulty for embedded systems that cannot be updated regularly. A survey by Aparcana-Tasayco et al. points out that most machine learning models are not validated to be resilient to dynamic, real-world scenarios, although they demonstrate high accuracy in their normal benchmarks \cite{aparcana2025systematic}. Consequently, a major challenge is that embedded anomaly detection should be evaluated using multiple criteria and should not only focus on accuracy but also on latency, memory footprint, and adaptability.

Through this brief survey, we cut through the complexity by offering a feasible, hardware-concerned perspective on the most plausible anomaly detection strategies for embedded systems. We bring out the trade-offs of most interest to the practitioners: what algorithms will run on either a Cortex-M microcontroller or a more powerful edge CPU, how they manage this changing nature of normal operation, and what can be used to fairly benchmark them. Our goal is to offer a strategic roadmap for engineers and other researchers as they find their way in the fast-paced intersection of TinyML, edge AI, and real-time security.

The remainder of this paper is organized as follows. Section \ref{s1} surveys the four dominant families of embedded anomaly detection methods: tree-based models, one-class learning approaches, lightweight neural networks, and statistical or threshold-based techniques. We analyze each in terms of detection performance, memory footprint, latency, and suitability for hardware platforms ranging from microcontrollers to edge CPUs. Section \ref{s2} discusses key trade-offs and emerging hybrid designs, while Section \ref{s3} identifies critical research gaps, particularly around concept drift, benchmarking standards, and adversarial robustness. Finally, Section \ref{s4} outlines promising directions for future work, and Section \ref{s5} concludes with practical guidance for deploying anomaly detection in real-world, resource-constrained environments.
\section{Embedded Anomaly Detection Key Approaches} \label{s1}
Anomaly detection on embedded and edge devices must balance detection performance with strict constraints on memory, latency, and power. In recent works, research has converged around four dominant methodological families, each offering distinct trade-offs between model complexity, adaptability, and hardware feasibility.

\subsection{Tree-Based Methods}
The class of ensembles based on trees, specifically Isolation Forest (IF), continues to be one of the most popular anomaly detectors used to date in embedded IoT devices because of their linear time complexity, small memory footprint, and data-scale insensitivity. IF separates anomalies by randomly dividing the features with the knowledge that anomalies take fewer splits in isolation. 

Recent literature proves that IF is still among the most efficient tree-related solutions in order to detect anomalies in embedded and IoT systems. Vasiljevic et al. suggested federated variant FLiForest, which can optimize IF on MicroPython-based edge devices, achieving an accuracy of more than 96 percent with a memory consumption of less than 160 KB, and emphasized that it fits well in resource-constrained and privacy-sensitive environments \cite{vasiljevic2025federated}. Another paper by Zahoor et al. \cite{zahoor2025robust} compared the IF with One-Class SVM and a hybrid CSAD technique in IoT security application that the IF is very robust and computationally efficient, although slightly worse than OCSVM. Chua et al. have demonstrated the performance of IF to detect web traffic anomalies with high accuracy and precision through structured data preprocessing and feature engineering \cite{chua2024web}. Besides, other researchers in \cite{cao2025anomaly} gave a comprehensive overview of isolation-based algorithms with low complexity, scalability, and robustness of IF, along with its effective support of streaming and distributed edge conditions.

\subsection{One-Class Learning Methods}
One-class learning models, such as One-Class SVM (OCSVM) and Support Vector Data Description (SVDD), are trained on a small boundary around normal data in a feature space. Although it was too memory-intensive historically to require microcontrollers, more recent computational economies (especially in model compression, linear approximations, and support vector reduction) have made systems using edge CPUs implemented feasible. The methods are best in cases where a normal behavior has a complicated, non-linear structure.

Recent studies have greatly contributed to improving one-class learning techniques of IoT and edge anomaly detection through better detection and computational performance. Katbi et al. suggested an enhanced interpolated Deep SVDD autoencoder with adversarial regularization, which improves the latent hypersphere representation to advance class separation in heterogeneous, high-dimensional data of IoT, exceeding the conventional shallow, as well as deep, baselines \cite{katbi2025one}. Ayad and his colleagues in another research paper \cite{ayad2024efficient} proposed a lightweight single-class detection system based on an asymmetric stacked autoencoder integrated with a profound neural network with detection rates of over 96\% and a high accuracy on IoT intrusion data with low inference time, which are suitable enough to be deployed on a real-time basis. The authors in this paper \cite{shi2024malware} examined one-class classification of IoT malware with TF-IDF and n-gram feature encoding, which demonstrates that they can use a model only trained with benign traffic and have a very high recall, as well as strong accuracy, which is remarkable in the context of changing threats. A paper by Yang et al. \cite{yang2021efficient} suggested an efficient OCSVM model that incorporates the Nyström method, Gaussian sketching, clustering, and Gaussian mixture models and allows using smaller memory and prediction time at a cost of quality per detector, which makes OCSVM more feasible in large-scale and constrained resource IoT applications. 

\subsection{Lightweight Neural Networks}
Small neural networks, including quantized autoencoders, 1D-CNNs, and pruned recurrent networks, are becoming more usable in resource-constrained devices of all sizes due to the TinyML revolution. These models identify anomalous events through reconstruction error or prediction variance, including those temporal dependencies that are typically overlooked by statistical or tree-based models.

The analysis and the current developments in lightweight neural network architectures have made it substantially more feasible to conduct deep anomaly detection on resource-limited edge and IoT devices. Sivapalan et al. proposed ANNet, a hybrid LSTM-MLP architecture, for open-source real-time ECG anomaly detector on wearable IoT sensors with about 97\% classification performance, which can significantly save energy through edge-level decision-making and gated wireless communication over ARM Cortex-M \cite{sivapalan2022annet}. Babalola and his colleaugues investigated AI-assisted edge cybersecurity through compact neural networks, including MobileNet, SqueezeNet, and TinyML models, noting that such neural networks are more effectively able to provide real-time, low-latency, anomaly detection services at small memory and power constraints \cite{babalola2024ai}. The authors in \cite{chang2025mematr} proposed MemATr, a memory-augmented lightweight transformer to detect video anomalies, which can attain competitive performance with just a small fraction of the number of parameters as a standard transformer model and latency to run of sub-50 ms on a mobile device. Amin et al. have shown that LSTM-autoencoder-based anomaly detection is practical in a microcontroller device, and quantization and TensorFlow Lite can be used to implement real-time monitoring of industrial equipment in real-time on the Arduino-class device \cite{amin2024real}. 

\subsection{Statistical and Threshold-Based Methods}
Statistical techniques such as standard deviation thresholds, moving averages, control charts, and lightweight PCA are the most straightforward and deterministic techniques of embedded anomaly detection. They need no training, provide constant time inference, and can run on footprints of less than 10 KB of memory. They are not as capable of finding complicated patterns as they would be with more advanced circuit-breaking techniques, but in safety-critical or ultra-low-power applications (e.g., a medical implant, an industrial alarm), they are peerless.

Recent research proves that statistical and threshold-based anomaly detection techniques are still quite useful in real-time, safety-critical, as well as low-latency industry settings. An et al. presented a constantly changing, online statistical log anomaly detection framework about the AIOps, which can be adaptable to real-time and automatically mitigates the data contamination, where improvements are up to 60\% higher F1-score than the conventional static pipelines \cite{an2022real}. The authors in \cite{huang2018evaluation} compared the traditional statistical and multivariate tools for wear monitoring with force sensor streams in CNC machining and found that they have a high robustness to parameter changes and (good) predictive performance in the case of progressive failures. A comparative industrial study by Mikayilov and Gardashova \cite{mikayilov2025modern} reported that simple yet powerful industrial lightweight unsupervised machine learning methods, such as IF and autoencoder baselines, give robust performance in noisy and high-dimensional manufacturing data, though smaller statistical models can be deployed much faster and improve operations quantitatively. 

\subsection{Comparative Summary}
Table~\ref{tab:embedded_ad_models} summarizes the key characteristics, performance, advantages, and limitations of the four main embedded anomaly detection approaches discussed in this section, based on recent literature.

\begin{table}[t]
\centering
\caption{Comparative Analysis of Embedded Anomaly Detection Approaches}
\label{tab:embedded_ad_models}
\scriptsize
\setlength{\tabcolsep}{4pt} % reduce horizontal padding
\renewcommand{\arraystretch}{1.1}
\begin{tabular}{|p{1.3cm}|p{1.6cm}|p{1.5cm}|p{1.5cm}|p{1.5cm}|}
\hline
\textbf{Approach + Ref} & \textbf{Description} & \textbf{Performance} & \textbf{Advantages} & \textbf{Limitations} \\
\hline

\multicolumn{5}{|c|}{\textbf{Tree-Based Methods}} \\
\hline
Isolation Forest \cite{vasiljevic2025federated, zahoor2025robust, chua2024web, cao2025anomaly} &
Unsupervised ensemble using random partitioning; anomalies isolated in fewer splits. Federated variants enable privacy-preserving edge learning. &
High accuracy/precision; low latency; microcontroller-efficient. &
Low RAM, linear-time, no labels, scalable. &
Struggles with temporal or clustered anomalies. \\
\hline

\multicolumn{5}{|c|}{\textbf{One-Class Learning}} \\
\hline
OCSVM / Deep SVDD / AE \cite{katbi2025one, ayad2024efficient, shi2024malware, yang2021efficient} &
Models learn boundary around normal data (kernel, hypersphere, or reconstruction). &
High recall \& precision; robust to zero-day attacks. &
Handles non-linear patterns; generalizes from benign-only data. &
High memory (OCSVM); sensitive to thresholds; complex training. \\
\hline

\multicolumn{5}{|c|}{\textbf{Lightweight Neural Networks}} \\
\hline
LSTM-AE, CNN, Transformer \cite{sivapalan2022annet, babalola2024ai, chang2025mematr, amin2024real} &
Quantized/pruned DNNs (LSTM, CNN, Transformer) deployed via TF Lite or MicroPython. &
High accuracy on time-series/multimodal data. &
Captures temporal dynamics; high detection quality. &
Higher RAM than trees; needs quantization \& feature engineering. \\
\hline

\multicolumn{5}{|c|}{\textbf{Statistical \& Threshold-Based}} \\
\hline
SPC, Moving Stats, RSM \cite{an2022real, huang2018evaluation, mikayilov2025modern} &
Rule-based: control charts, deviation thresholds, online statistical tests. &
Near-zero latency; robust to parameter shifts; improves F1-score. &
<10 KB RAM; deterministic; no training; safety-critical ready. &
Misses complex anomalies; assumes stationarity; poor on non-Gaussian data. \\
\hline
\end{tabular}
\end{table}

\section{Discussion}\label{s3}
The comparative landscape of embedded anomaly detection reveals a clear trade-off between model expressiveness and deployment feasibility. Tree-based methods like IF consistently emerge as the default choice for ultra-constrained microcontrollers due to their minimal memory footprint, linear-time inference, and robustness to high-dimensional sensor data. However, their inability to model temporal dynamics limits their effectiveness in sequential monitoring tasks such as vibration analysis or ECG streams. In contrast, lightweight neural networks, particularly quantized LSTM-autoencoders and compact CNNs, demonstrate superior accuracy in time-series contexts by capturing contextual dependencies, but at the cost of increased memory and computational demands that often exceed the capabilities of Cortex-M-class devices. One-class classification methods, while theoretically powerful for modeling complex decision boundaries in normal behavior, remain largely impractical for edge deployment due to their reliance on support vectors or dense kernel computations that scale poorly with data volume. Statistical and threshold-based approaches, though simplistic, retain relevance in safety-critical applications where deterministic latency and interpretability outweigh detection sophistication. Notably, the most effective real-world systems increasingly adopt hybrid strategies: using a lightweight unsupervised filter (e.g., IF) for initial anomaly screening, followed by a more resource-intensive model only when triggered. This cascaded architecture balances responsiveness with precision while conserving power, critical for battery-operated IoT nodes. Despite advances in model compression and TinyML toolchains, the gap between research prototypes and production-ready edge deployment remains wide, primarily due to inconsistent evaluation practices that prioritize accuracy over real-world constraints like latency, energy, and concept drift resilience.

\section{Research Gaps}\label{s2}
Current literature exhibits several critical shortcomings that hinder the practical adoption of embedded anomaly detection. First, there is a lack of standardized benchmarks that jointly evaluate detection performance alongside hardware-aware metrics such as inference latency, RAM usage, power consumption, and model update overhead. Most studies report accuracy or F1-score in isolation, often on static offline datasets, which fails to reflect the dynamic, streaming nature of real sensor data. Second, the challenge of concept drift, where the definition of “normal” evolves due to aging hardware, environmental changes, or software updates, is rarely addressed with viable on-device adaptation mechanisms. Existing methods either assume stationarity or require cloud-assisted retraining, breaking the premise of true edge autonomy. Third, the evaluation of adversarial robustness is almost entirely absent; lightweight models are highly susceptible to input perturbations, yet few works consider security-efficacy trade-offs. Fourth, there is minimal investigation into cross-platform portability: a model optimized for TensorFlow Lite Micro on an ARM Cortex-M4 may not translate efficiently to ESP32 or RISC-V architectures without significant re-engineering. Finally, the human-in-the-loop aspect, such as explainability for field engineers or configurable false-positive tolerance,is overlooked, reducing trust in autonomous anomaly alerts in industrial settings.

\section{Future Work}\label{s4}
To bridge these gaps, future research should prioritize the development of holistic, hardware-aware evaluation frameworks that enforce joint optimization of accuracy, latency, memory, and energy, ideally integrated into community-driven initiatives like MLPerf Tiny. Online learning techniques must be co-designed with anomaly detectors to enable continuous adaptation to concept drift without catastrophic forgetting or excessive computational overhead; this includes exploring incremental PCA, streaming autoencoders, or federated one-class models that update only split thresholds or reconstruction baselines. Bio-inspired optimization approaches could provide new adaptation paradigms. Just as bacterial foraging optimization algorithms mimic E. coli behavior \cite{makhlouf2024enhanced}, similar mechanisms could enable anomaly detectors to dynamically adjust decision boundaries while navigating concept drift with minimal energy expenditure. Hardware-software co-design should be advanced through tighter integration of neural architecture search (NAS) with embedded constraints, yielding models that are natively quantized, sparsified, and memory-optimized during training. Hybrid optimization frameworks offer promising directions. Techniques that integrate particle swarm optimization with artificial potential fields demonstrate how continuous recalculation of optimal paths, while dynamically adjusting to environmental changes and avoiding obstacles during replanning, can reduce computation time while maintaining efficiency \cite{benmachiche2025adaptive}. Such approaches could inspire energy-aware model updating strategies that dynamically adjust anomaly thresholds with minimal overhead. Additionally, research into lightweight adversarial training or input sanitization layers could enhance robustness while preserving efficiency. Temporal adaptation mechanisms represent another frontier. Architectures combining Long Short-Term Memory networks with specialized attention processors can highlight important behavioral patterns over time while filtering irrelevant data \cite{sedraoui2025lstm}. Crucially, unlike conventional systems that rely on fixed features, such models can adapt dynamically to new and evolving anomaly patterns, providing resilience against emerging threats without cloud dependency. Standardized deployment pipelines, supporting seamless conversion from PyTorch or TensorFlow to MCU-compatible formats with automatic memory budgeting, would accelerate real-world adoption. 
Attention mechanisms and transformer architectures represent promising directions for next-generation embedded anomaly detection. Recent advances in sparse attention frameworks \cite{boufaida2025tsa} demonstrate how computational efficiency can be maintained while preserving contextual awareness in sequential data. Similarly, hybrid CNN-Transformer models \cite{boutabia2025hybrid} illustrate the potential for combining local feature extraction with global relationship modeling. Future research should investigate hardware-aware implementations of these architectures that maintain their expressive power while respecting the stringent memory and latency constraints of edge devices.
Finally, human-centered design principles must be incorporated: anomaly scores should be interpretable (e.g., via feature attribution on microcontrollers), and systems should support operator feedback loops to refine thresholds or label edge cases, enabling semi-supervised lifelong learning directly on the device.
\section{Conclusion}\label{s5}
This survey has looked over the terrain of real-time anomaly detection in embedded and edge IoT systems, focusing on techniques that balance the detection effectiveness and stringent hardware limitations. Approaches based on trees, specifically IFs, provide a viable option that can be deployed on ultra-constrained microcontrollers, can have low latency (less than 50 ms), low memory footprint (less than 160 KB), and achieve high performance without labeled data. Deep autoencoders and optimized OCSVM implementations, which use a single class, are more accurate and resistant to more complicated patterns, but require more resources and can be used on edge hardware such as Raspberry Pi or Jetson Nano. The lightweight neural networks (e.g., quantized LSTM-AEs) are intermediate solutions, which are capable of temporal dynamics in time-series data but still can be deployed on TinyML toolchains. Conversely, statistical and threshold-based approaches, although basic, are still useful in safety-critical or deterministic real-world environments where human readability and zero-training execution are more important than detection advanced detection capabilities.

Most importantly, there is no single set of approaches that is optimal in all situations. The best selection is based on the hardware to target, data modality, the presence of labeled anomalies, and real-time issues. The subsequent stage in co-design is the evolution of online learning with concept drift, benchmark standardization with latency and energy metrics, and explainability features to create operator trust. Lightweight, adaptive, and hardware-aware anomaly detection will continue to be a vital component of ensuring the intelligent edge as the IoT deployments become increasingly large and critical.

\bibliography{biblio}
\bibliographystyle{plain}

\end{document}